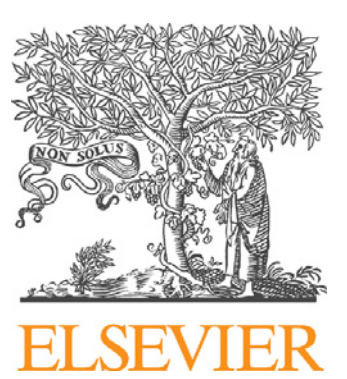

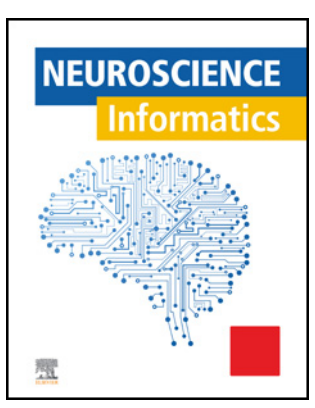

Original article

# Modified Weibull distribution for Biomedical signals denoising

A.M. Adam [a,*], B.S. El-Desouky [a], R.M. Farouk [b]

[a] Department of Mathematics, Faculty of Science, Mansoura University, Mansoura, Egypt, P.O. Box, Egypt
[b] Department of Mathematics, Faculty of Science, Zagazig University, Zagazig, Egypt, P.O. Box, Egypt

ABSTRACT

A wide range of signs are acquired from the human body called Biomedical signs or biosignals, they can be at the cell level, organ level, or sub-atomic level. Electroencephalogram is the electrical activity from the cerebrum, the electrocardiogram is the electrical activity from the heart, electrical action from the muscle sound signals referred to as electromyogram, the electroretinogram from the eye, and so on. Studying these signals can be so helpful for doctors, it can help them examine and predict and cure many diseases.
However, Biomedical signals are often affected by various types of noise, it's important to denoise the signals to get accurate information from them, the denoising process is solved by proposing an entirely novel family of flexible score functions for blind source separation, based on a family of modified Weibull densities. To blindly extract the independent source signals, we resort to the popular Fast independent component analysis approach, to adaptively estimate the parameters of such score functions, we use an efficient method based on maximum likelihood. The results obtained using modified Weibull densities in our technique are better than those obtained by other distribution functions.

## 1. Introduction

Blind source separation (BSS) is a high-level image/signal processing technique and has numerous applications such as sound signals, communication, images, and biomedicine [1–4]. BSS aims to retrieve the source (images/signals) from a noised source with little known information. Various BSS algorithms have been discussed from various points of view, including non-Gaussianity [5], mutual information minimization [6], maximum likelihood [7], tensors [8], principle component analysis (PCA) [9], and neural networks [10–12]. Regarding BSS, denoising and optimization methods play the most important roles. The noise separation step measures the separability, and the optimization step is used to get the optimum solution for the objective function which we get from the denoising mechanism. Generalized distributions usually give good results of blind denoising due to the variant properties of its sub-models. In the independent component analysis (ICA) framework, accurately estimates the statistical model of the sources is still an open and challenging problem [2]. Practical BSS scenarios employ difficult source distributions and even situations where numerous sources with variant probability density functions (pdf) mixed together. Towards this direction, many parametric density models have been made available in recent literature. Such models as, the generalized Gaussian density [13], the generalized gamma density [14], and even combinations and generalizations such as super and generalized Gaussian mixture model [15], the Pearson family of distributions [16], the generalized alfa-beta distribution (AB-divergences) [17] and even the so-called extended generalized lambda distribution [18] which is an extended parameterization of the aforementioned generalized lambda distribution and generalized beta distribution models [19].

Although FastICA has some disadvantages, as it often leads to local minimum solutions due to the difficulty of optimizing the log-likelihood function, which means the suitable source signals are not isolated, and also the order of the independent components (ICs) is difficult to be determined, but FastICA still one of the most powerful techniques and usually drive very good results.

However, studying medical signals became very important and essential; it is very difficult to get useful information from these signals directly in the time domain just by observing them. They are basically non-linear and non-stationary in nature. Biomedical signals are usually affected by various types of noise, which is considered a challenging problem, for example, one of the challenges of electroencephalogram (EEG) technology is that electrical activity generated by the brain is minuscule, on the order of a millionth of a volt. Consequently, scalp recorded electrical activity consists of a

* Corresponding author.
*E-mail addresses:* amer.adam@hotmail.com (A.M. Adam), b_desouky@yahoo.com (B.S. El-Desouky), rmfarouk1@yahoo.com (R.M. Farouk).

**Table 1**
The MWD sub-models.

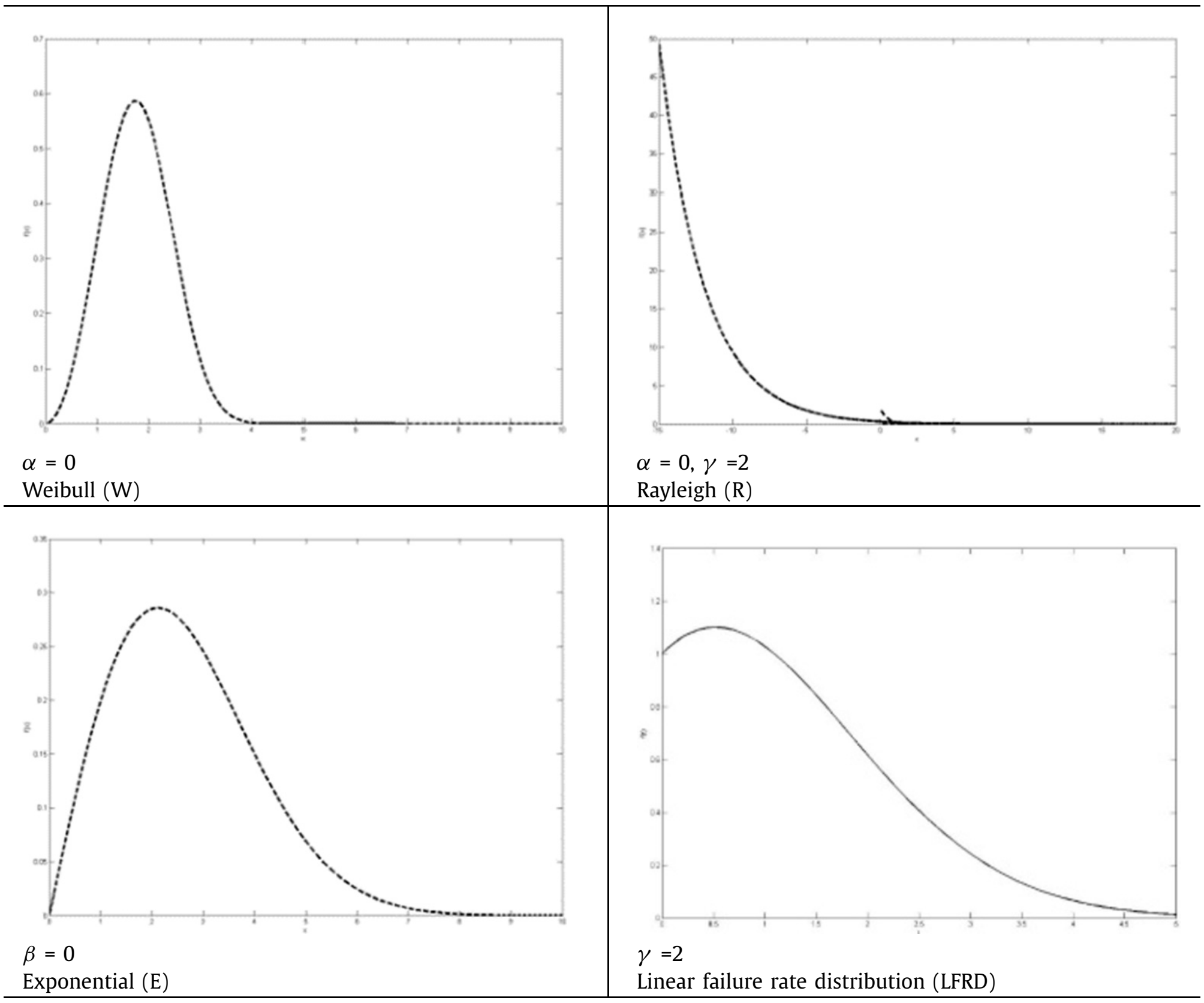

| | |
|---|---|
| $\alpha$ = 0<br>Weibull (W) | $\alpha$ = 0, $\gamma$ =2<br>Rayleigh (R) |
| $\beta$ = 0<br>Exponential (E) | $\gamma$ =2<br>Linear failure rate distribution (LFRD) |

The MWD sub-models, shows the specific values of the parameters used to generate the above mentioned four special cases.

mix of genuine brain signals combined with lots of noise - termed artifact - generated by other parts of the body, such as heart activity, eye movements, and blinks, other facial muscle movements, etc., which produce electrical signals about 100 times greater than those produced by the brain. Also, the general background noise comes from outside the brain.

Hence, in the need of extracting important information from the signals, noise has to be removed. To achieve that, numerous advanced signal processing techniques have been developed. In this paper, we present the Modified Weibull distribution (MWD) with ICA to remove noise from biomedical signals. We evaluated the accuracy of the proposed algorithm, the numerical results, show that the MWD gives very good results. We organized the rest of this paper as follows: Section 2, we present the BSS model Section 3, we present Independent Component Analysis (ICA). Section 4, we discuss the MWD. Finally, we present the computational efficient performance of the proposed technique.

## 2. Blind source separation (BSS) model

Let $S(t) = [s_1(t), s_2(t), \ldots, s_N(t)]^T (t = 1, 2, \ldots, l)$ denote an independent source signal vector that comes from N signal sources, then we can get the observed mixtures

$$X(t) = [x_1(t), x_2(t), \ldots, x_K(t)]^T, (N = K)$$

under the circumstances of the instantaneous linear mixture. This leads us to the BSS model

$$X(t) = AS(t) \tag{1}$$

where A is $N \times N$ mixing matrix. The target of the BSS algorithm is to recover the sources from mixtures x(t) by using

$$U(t) = WX(t) \tag{2}$$

where W is $N \times N$ separation matrix and $U(t) = [u_1(t), u_2(t), \ldots, u_N(t)]^T$ is the estimate of N sources.

Usually, sources are assumed to be unit-variance and zero-mean signals including at most one having a Gaussian distribution. To solve the source estimation problem, the unmixing matrix W must be determined. Generally, the majority of BSS approaches perform ICA, by essentially optimizing the negative log-likelihood (objective) function concerning the un-mixing matrix W such that

$$L(u, W) = \sum_{l=1}^{N} E[\log p_{ul}(u_l)] - \log|\det(W)| \tag{3}$$

Where E[.] represents the expectation operator and $p_{u1}(u_1)$ is the model for the marginal pdf of $u_l$, for all $l = 1, 2, \ldots, N$. In effect, when correctly hypothesizing upon the distribution of the sources, the maximum likelihood (ML) principle leads to estimating functions, which are the score functions of the sources [15]

$$\varphi_l(u_l) = -\frac{d}{du_l} \log p_{ul}(u_l) \tag{4}$$

In principle, the separation criterion can be optimized by any suitable ICA algorithm where contrasts are utilized (see; e.g., [2]). The FastICA [3], based on

$$W_{k+1} = W_k + D(E[\varphi(u)u^T] - \mathrm{diag}(E[\varphi_l(u_l)u_l]))W_k \tag{5}$$

where, as defined in [4],

$$D = \mathrm{diag}\left(\frac{1}{E[\varphi_l(u_l)u_l] - E[\varphi_l'(u_l)]}\right) \tag{6}$$

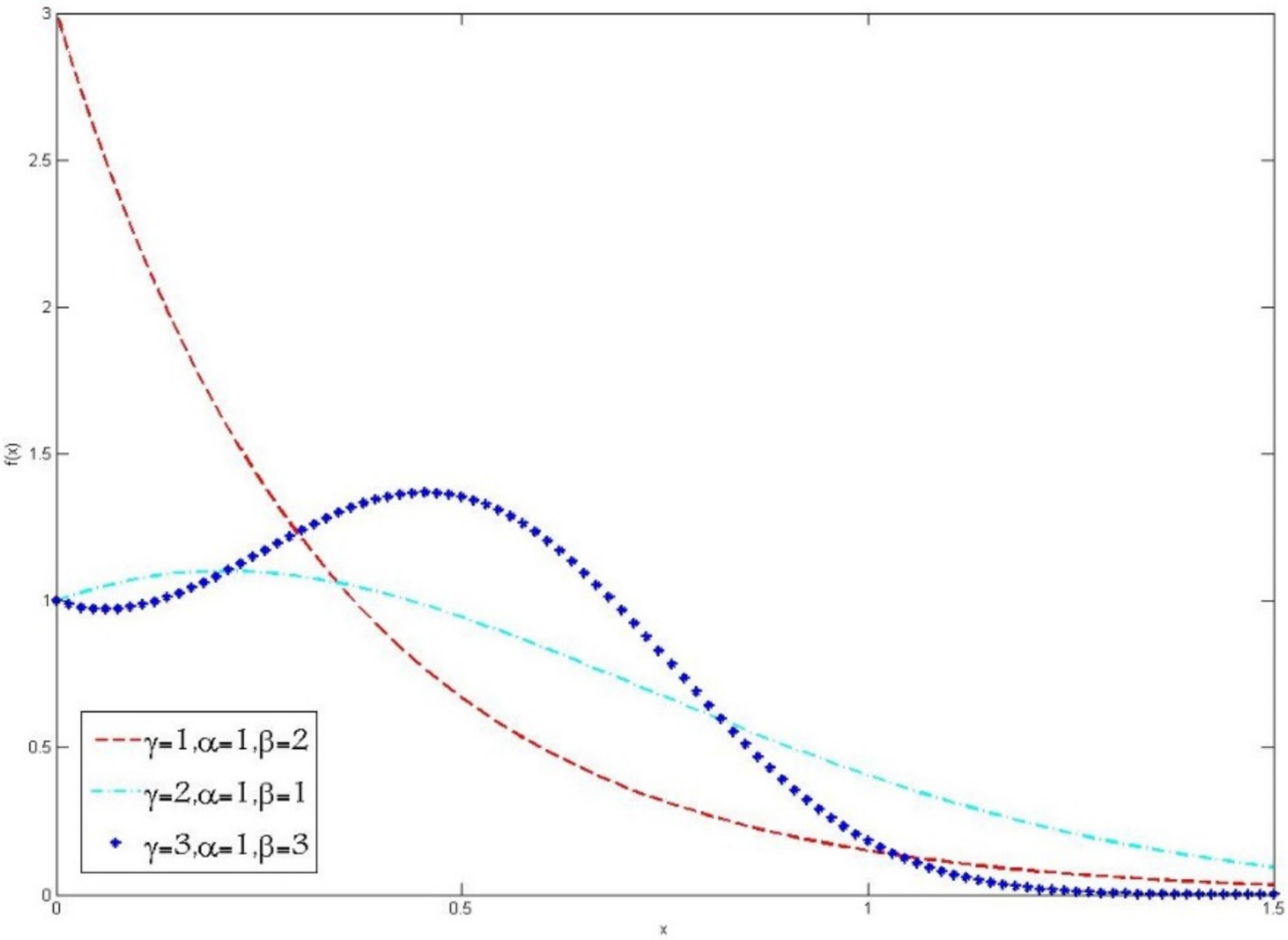

**Fig. 1.** The MWD with fixed $\alpha$ =1.

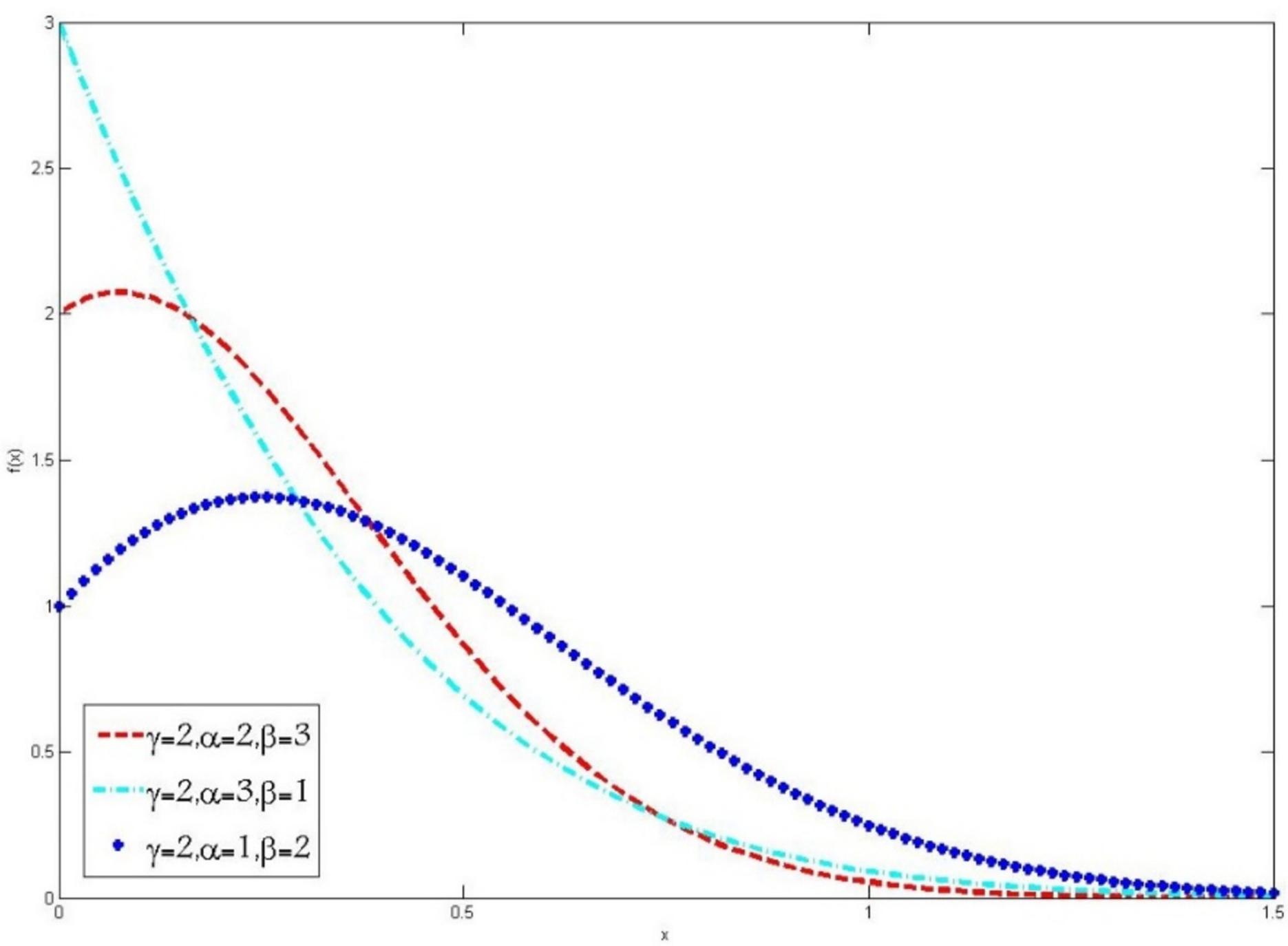

**Fig. 2.** The MWD with fixed $\gamma$ =2.

**Table 2**
Parameter estimation by using GA.

| $\alpha$ | $\beta$ | $\gamma$ | $\hat{\alpha}$ | $\hat{\beta}$ | $\hat{\gamma}$ | Err |
|---|---|---|---|---|---|---|
| 3 | 4 | 2 | 2.97 | 4.11 | 1.86 | 0.02 |
| 5.2 | 6.8 | 2.5 | 5.27 | 6.80 | 2.42 | 0.06 |
| 1.9 | 8.2 | 5.7 | 1.98 | 8.12 | 563 | 0.006 |

where $\varphi(\mathrm{t}) = [\varphi_1(\mathrm{u}_1), \varphi_2(\mathrm{u}_2), \ldots, \varphi_\mathrm{n}(\mathrm{u}_\mathrm{n})]^\mathrm{T}$, valid for all $\mathrm{l} = 1, 2, \ldots, \mathrm{n}$.

In the following section, we propose MWD for signal modeling.

## 3. Independent Component Analysis (ICA)

### 3.1. Definition of ICA

"It's a method for finding underlying factors or components from multivariate (multi-dimensional) statistical data. What distinguishes the ICA from other methods is that it looks for components that are both statistically independent, and non-Gaussian." [20]

Now, assume that we observe n linear mixtures $\mathrm{x}_1, \ldots, \mathrm{x}_\mathrm{n}$ of n independent components [21]

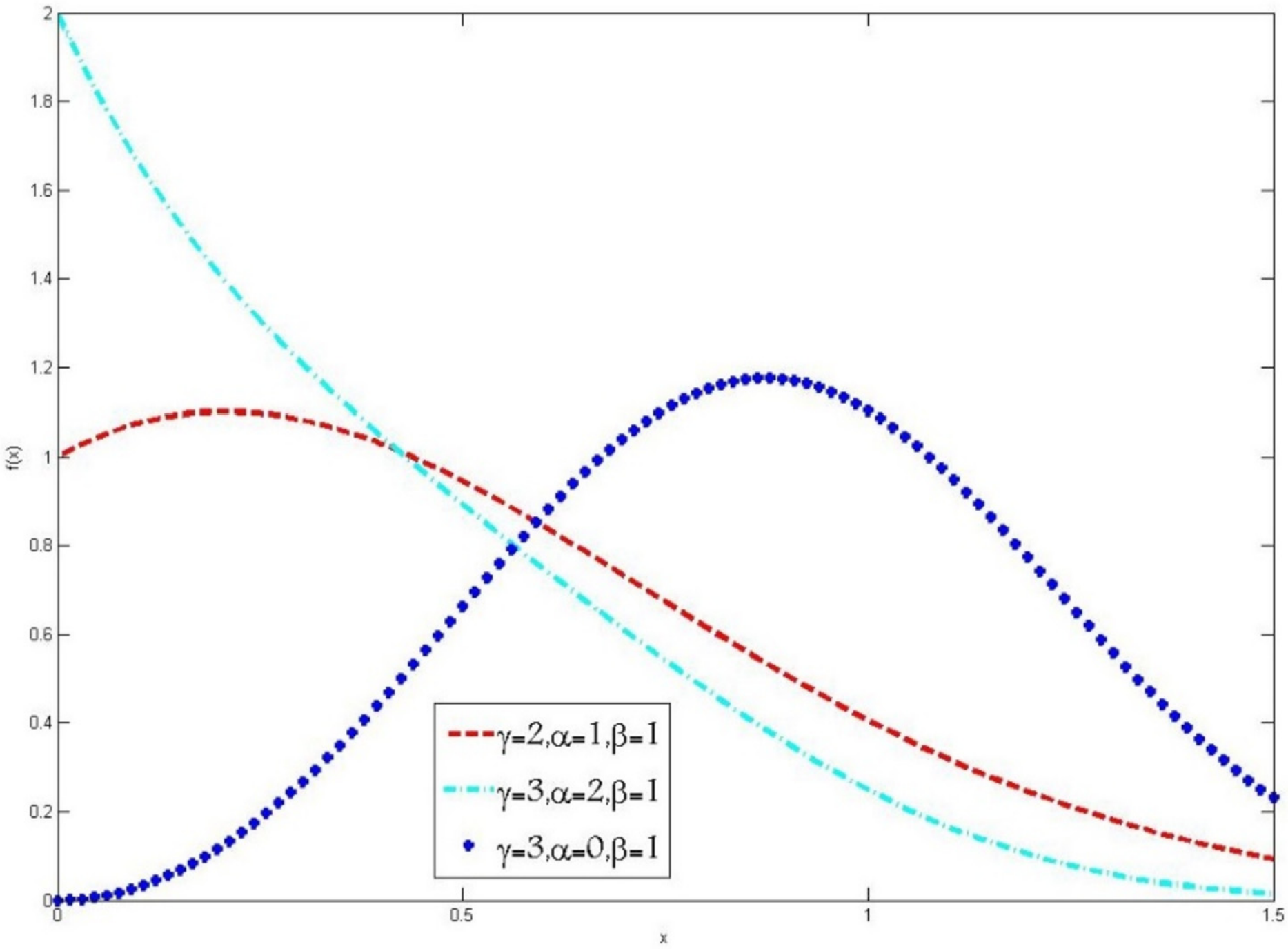

**Fig. 3.** The MWD with fixed $\beta$ =1.

**Table 3**
The performance of the proposed denoising algorithm for EEG signals.

| Signal | Mean Squared Error (MSE) | Mean Absolute Error (MAE) | Signal To Noise Ratio (SNR) | Peak Signal To Noise Ratio (PSNR) | Cross Correlation (CC) |
|---|---|---|---|---|---|
| EEG 1 | 0.1696 | 0.4028 | 7.7473 | 19.3231 | 0.9962 |
| EEG 2 | 0.1707 | 0.4049 | 7.6990 | 16.3919 | 0.9966 |

$$x_j = a_{j1}s_1 + a_{j2}s_2 + \ldots + a_{jn}s_n, \ for\ all\ j \tag{7}$$

The time index t has been dropped; in the ICA model [20][21], it is assumed that each mixture $x_j$ and each independent component $s_k$ is a random variable, instead of a proper time signal. The observed values $x_j(t)$, e.g., the microphone signals, are then a sample of this random variable. As a preprocess to simplify the calculation, we can assume that both the mixture variables and the independent components have zero mean: If not, then the observed variables xi can always be centered by subtracting the sample mean, this makes the model zero-mean. It would be convenient to use a vector-matrix notation instead of the sums like in the previous equation. Let's denote by x the random vector whose elements are the mixtures $x_1, \ldots, x_n$, and by s the random vector with elements $s_1, \ldots, s_n$, and by A the matrix with elements $a_{ij.}$ The above mixing model can be written as

$$\boldsymbol{x} = \boldsymbol{As} \tag{8}$$

Also, the model can be written as

$$\boldsymbol{x} = \sum_{i=1}^{n} \boldsymbol{a}i\ si \tag{9}$$

The statistical model in Eq. (9) is called the ICA model.

It is a generative model; it describes how the observed data are generated by a process of mixing the components si.

The key idea for ICA is very simple, assume that the components si are statistically independent. Also, they must have non-Gaussian distributions.

### 3.2. The FastICA algorithm

We introduced different measures of non-Gaussianity [20][21], i.e. objective functions for ICA estimation. In practice, also we need an algorithm for maximizing the contrast function, one of the most efficient algorithms of the ICA is the FastICA Algorithm, and this is what we will use in our new proposed method.

## 4. Proposed algorithm

### 4.1. Modified Weibull distribution (MWD)

Following [22] MWD is a new Modification of the two parameters Weibull distribution, and also it has been extended by other authors as a special sub-model as in [23] and [24].

The pdf of MWD or, modified Weibull probability density (MWPD) is defined as:

$$f\ (x) = \left(\propto + \beta\gamma x^{\gamma-1}\right) \times exp\left\{-\propto x - \beta x^{\gamma}\right\},\ x \geq 0 \tag{10}$$

where $\gamma > 0$, $\alpha, \beta \geq 0$ such that $\alpha + \beta > 0$. It is clear that the MWD is very flexible. This is so since many other distributions can be considered as special cases of MWD, by selecting the appropriate values of the parameters. These special cases include four distributions as shown in Table 1. In Figs. 1, 2, 3 there are the distributions generated from MWD by changing the parameters.

The corresponding cumulative distribution is given by:

$$F\ (x) = 1 - exp\left\{-\propto x - \beta x^{\gamma}\right\},\ x \geq 0 \tag{11}$$

Where $\gamma > 0$, $\alpha, \beta \geq 0$ such that $\alpha + \beta > 0$.

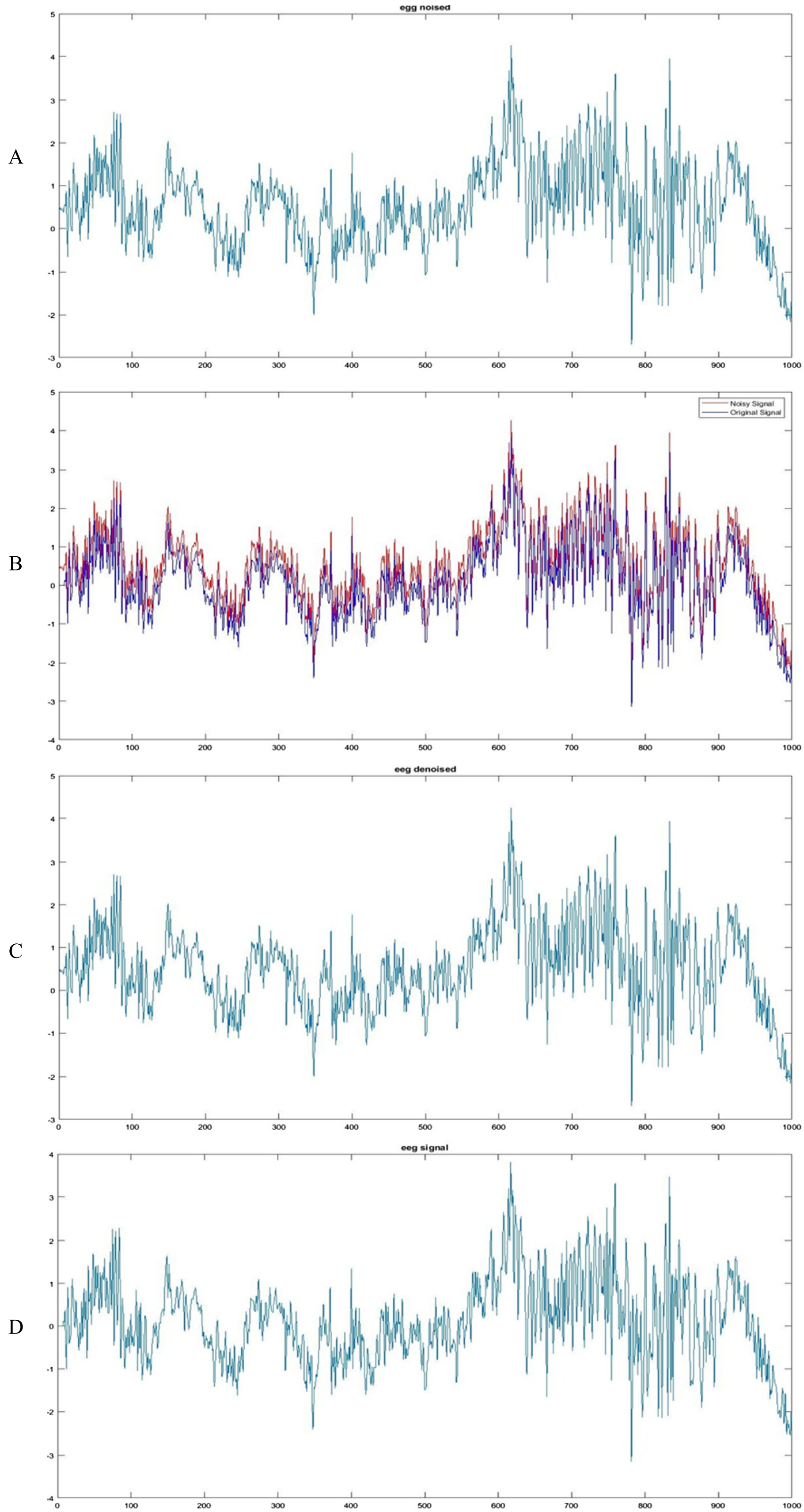

**Fig. 4.** (EEG signal 1): A noised signal, B noised signal (original signal in blue and noise in red), C denoised signal, D original signal.

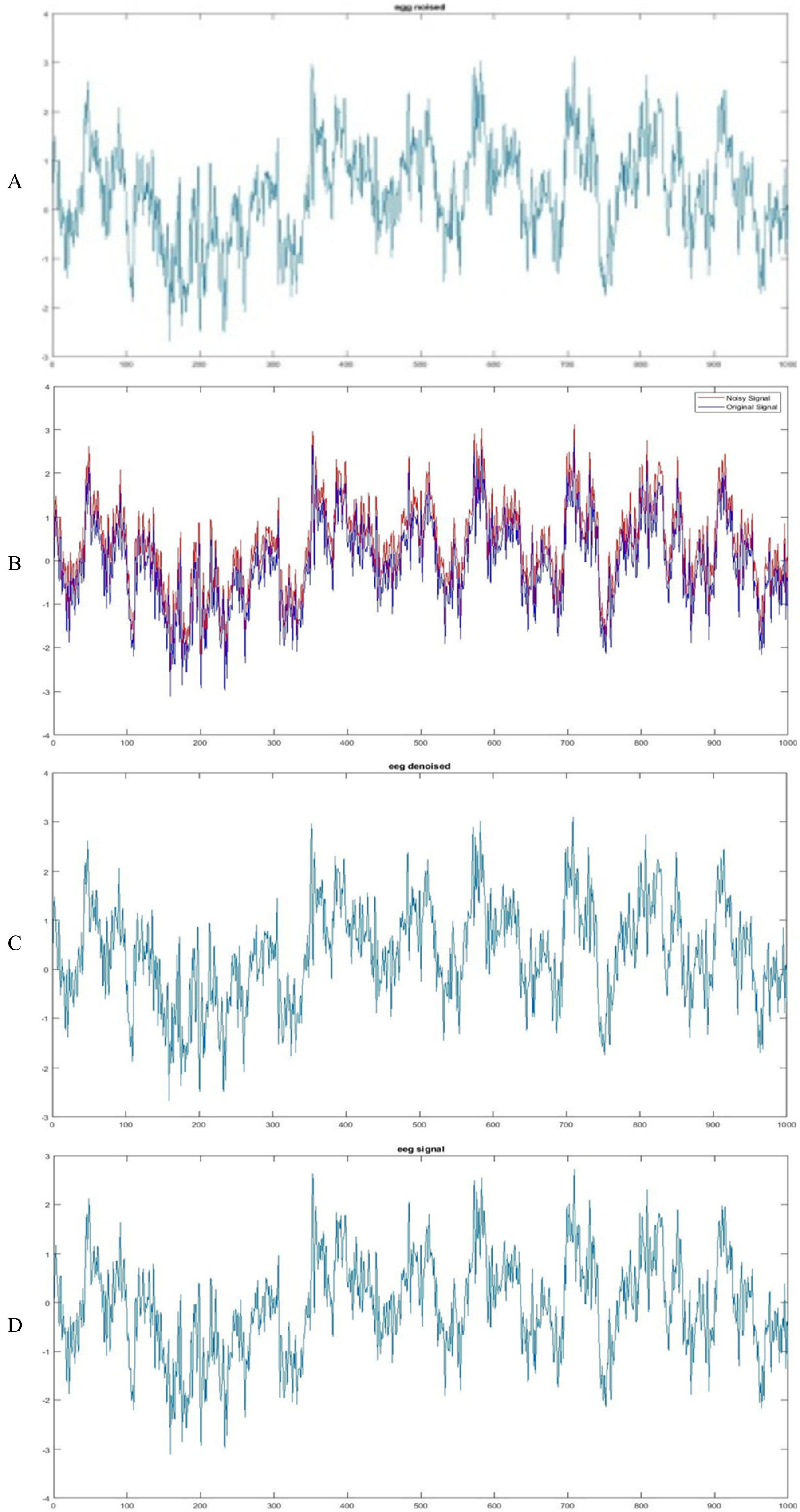

**Fig. 5.** (EEG signal 2): A noised signal, B noised signal (original signal in blue and noise in red), C denoised signal, D original signal.

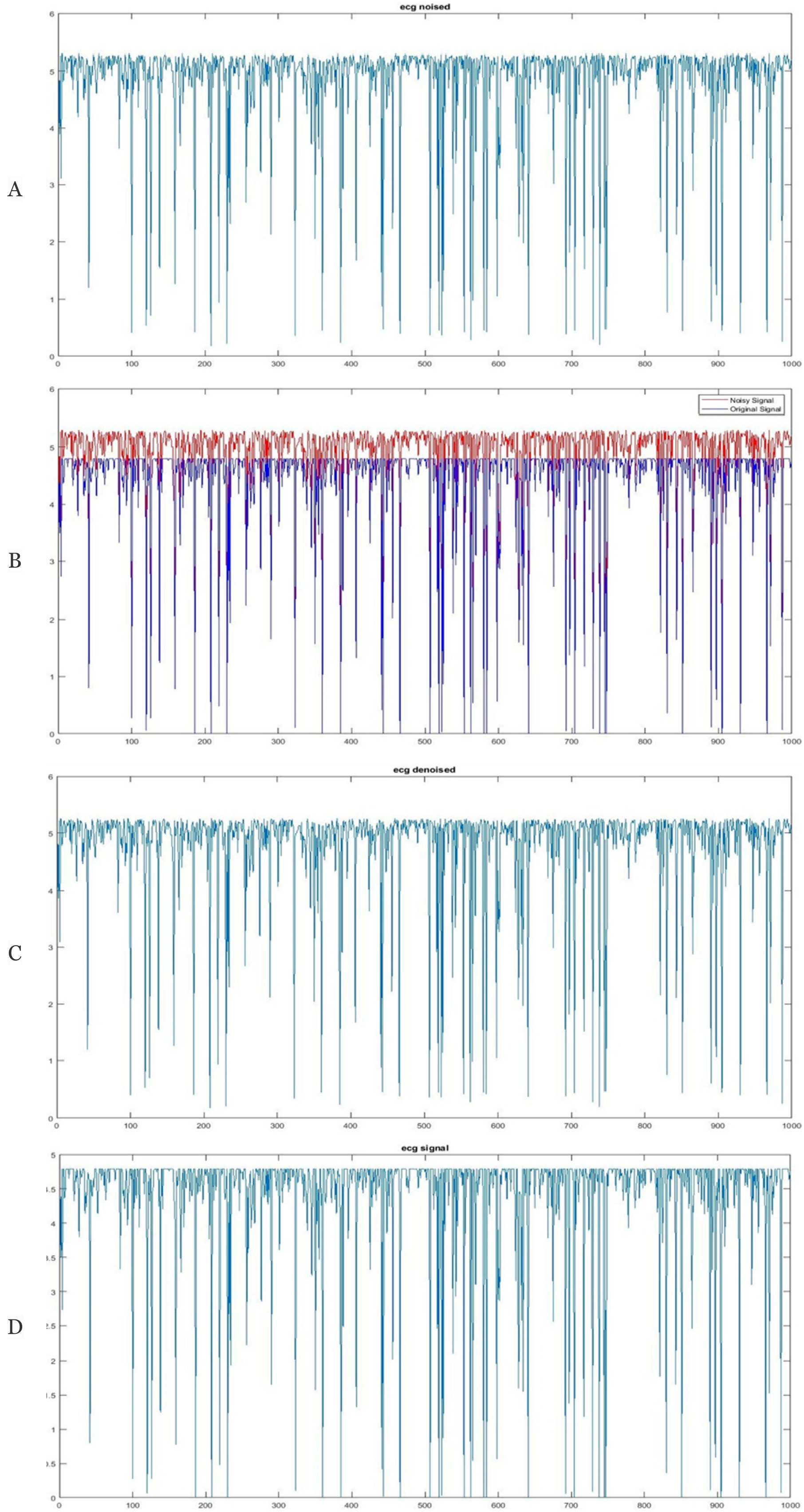

**Fig. 6.** (ECG signal 1): A noised signal, B noised signal (original signal in blue and noise in red), C denoised signal, D original signal.

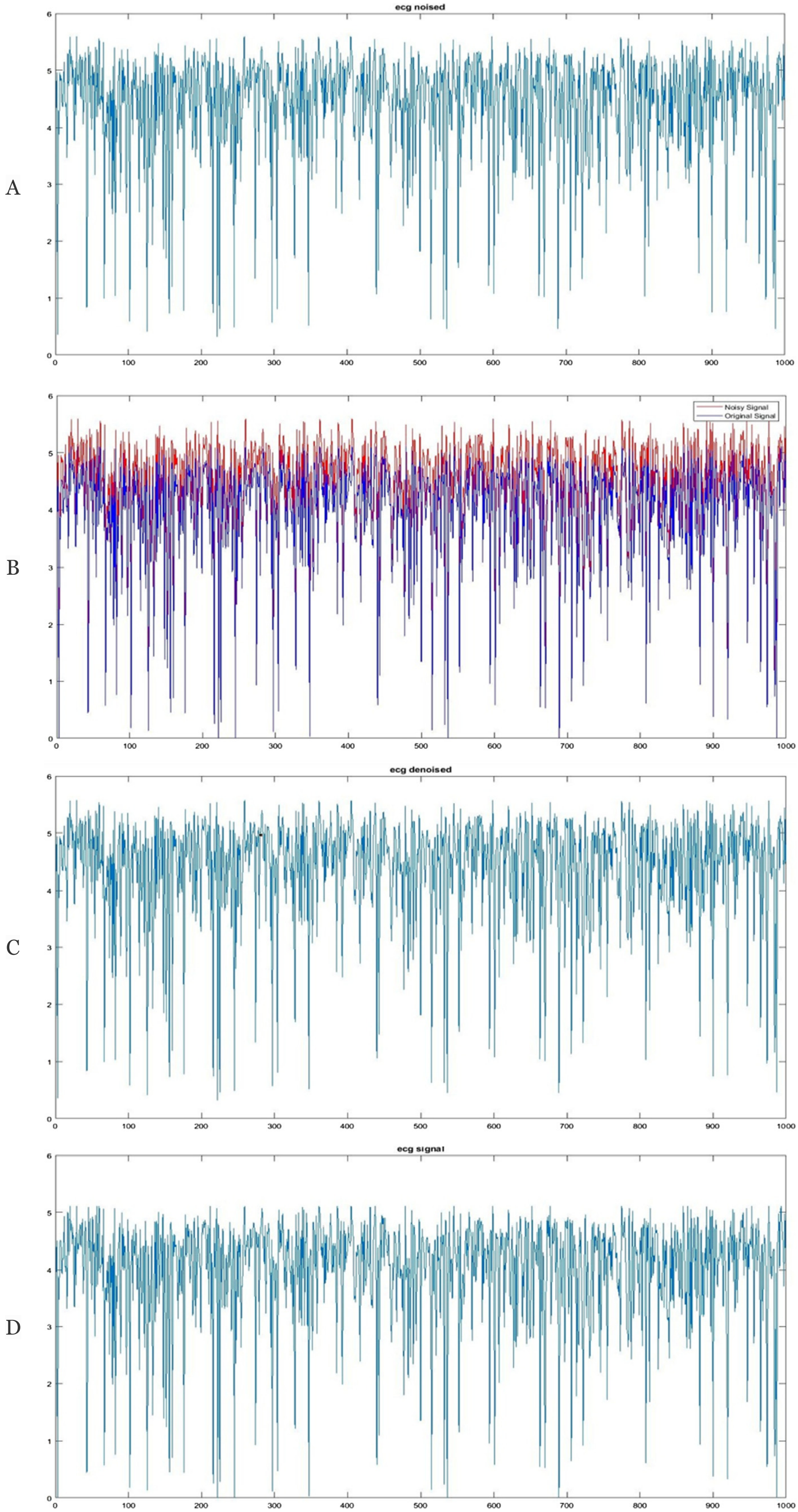

**Fig. 7.** (ECG signal 2): A noised signal, B noised signal (original signal in blue and noise in red), C denoised signal, D original signal.

**Table 4**
The performance of the proposed denoising algorithm for ECG signals.

| Signal | Mean Squared Error (MSE) | Mean Absolute Error (MAE) | Signal To Noise Ratio (SNR) | Peak Signal To Noise Ratio (PSNR) | Cross Correlation (CC) |
|---|---|---|---|---|---|
| ECG 1 | 0.1588 | 0.3895 | 21.0607 | 21.6062 | 0.9964 |
| ECG 2 | 0.1571 | 0.3876 | 20.3239 | 22.2007 | 0.9965 |

### 4.2. Maximum Likelihood Estimation (MLE) method

The MLE procedure employed is to determine the values of the MWD parameters, $\alpha$, $\beta$, and $\gamma$. to illustrate the method. The first-order optimality conditions below are used for this purpose.

### 4.3. Parameter estimation

To estimate the parameters of MWD, the maximum likelihood is used. Let $X_1, X_2 \ldots, X_n$ be a sample of size N from an MWD. Then the log-likelihood function ($\mathcal{L}$) is given by:

$$\mathcal{L} = log\,\ell = \sum_{i=1}^{n}\left[log\left(\propto + \beta\gamma x_i^{\gamma-1}\right)\right] + \sum_{i=1}^{n}\left(-\propto x_i - \beta x_i^{\gamma}\right) \quad (12)$$

Therefore, the maximum likelihood estimation of $\alpha$, $\beta$, and $\gamma$ are derived from the derivatives of $\mathcal{L}$. They should satisfy the following equations:

$$\frac{\partial \mathcal{L}}{\partial \alpha} = 0, \quad \frac{\partial \mathcal{L}}{\partial \beta} = 0, \quad \frac{\partial \mathcal{L}}{\partial \gamma} = 0$$

$$\frac{\partial \mathcal{L}}{\partial \alpha} = \sum_{i=1}^{n} \frac{1}{\propto + \beta\gamma x_i^{\gamma-1}} - n x_i \quad (13)$$

$$\frac{\partial \mathcal{L}}{\partial \beta} = \sum_{i=1}^{n} \frac{\gamma x_i^{\gamma-1}}{\propto + \beta\gamma x_i^{\gamma-1}} - \sum_{i=1}^{n} x_i^{\gamma} \quad (14)$$

$$\frac{\partial \mathcal{L}}{\partial \gamma} = \sum_{i=1}^{n} \frac{\beta x_i^{\gamma+1} + \beta\gamma x_i^{\gamma}\, log\, x_i}{\propto + \beta\gamma x_i^{\gamma-1}} - \sum_{i=1}^{n} \beta x_i^{\gamma}\, log\, x_i \quad (15)$$

To estimate the value of parameters, the system of equations (13, 14, 15) must be solved. However, it is difficult to solve this system so, the genetic algorithm (GA) [25][26] will be used as an alternative numerical method to estimate the parameters. The GA optimization technique lies in the fact that it can minimize the negative of the log-likelihood objective function in (3), essentially without depending on any derivative information.

## 5. Numerical results

Numerical experiments show that the GA method converges to an acceptably accurate solution with substantially fewer function evaluations. We generate random samples from modified Weibull distribution along with different combinations of the parameters, and then the ML estimates are obtained using the GA method as in Table 2.

In MATLAB, an implemented procedure called (ga) can be used to obtain the ML estimate, and such procedure is very fast and accurate. The Proposed algorithm represents excellent results for both EEG and electrocardiogram (ECG) signals.

## 6. Experimental results

We resolve to FastICA algorithm for (BSS). This algorithm depends on the estimated parameters and an un-mixing matrix W which is estimated by the FastICA algorithm, using real data set we used a data sample of size (1000). By substituting (10) into (4) for the source estimates $u_l$, $l = 1, 2, \ldots, n$, it quickly becomes clear that the proposed score function inherits a generalized parametric structure, which can be attributed to the highly flexible FWD parent model. So, a simple calculus yields the flexible BSS score function

$$\varphi_l(u_l) = -\frac{d}{du_l}\log\left[\left(\propto + \beta\gamma x^{\gamma-1}\right) \times \exp\left\{-\propto x - \beta x^{\gamma}\right\}\right] \quad (16)$$

In principle $\varphi_l(u_l \mid \theta)$ is capable of modeling a large number of signals as well as various other types of challenging heavy- and light-tailed distributions. Experiments were done to investigate the performance of our method through two applications (one for EEG signal denoising (using two different EEG signals) and the other for ECG signal denoising (using two different ECG signals) when Gaussian noise is presented.

**Example 1.** Electroencephalogram (EEG), electrical activity from the brain, one of the most important signals from the human body, studying it is very important to doctors who work in this field of medicine, monitoring and observing changes in these signals help them to cure and protect brain diseases. However, these signals may be corrupted due to various noising interferences. In this example we use the proposed algorithm for denoising two different EEG signals, the results are shown in Fig. 4 for EEG signal 1, and Fig. 5 for EEG signal 2. The performance of the proposed denoising algorithm is evaluated using: MSE, MAE, SNR, PSNR, and cross-correlation, shown in Table 3.

**Example 2.** Electroencephalogram (ECG) is an electrical activity from the heart, it is usually infected with various types of noise just as other types of biomedical signals. In this example we use the proposed algorithm for denoising two different ECG signals, the results are shown in Fig. 6 for ECG signal 1, and Fig. 7 for EEG signal 2, The performance of the proposed denoising algorithm is evaluated using: MSE, MAE, SNR, PSNR, and cross-correlation, shown in Table 4.

## 7. Conclusion

In this paper, we introduced a technique for biomedical signals denoising and blind Source separation based on modified Weibull distribution. Our proposed technique outperforms existing solutions in terms of denoising quality and computational cost. We applied our technique on EEG and ECG signals, and the results were excellent, the technique can be extended to be applied to several other biomedical signals.

As future work, we plan to use the algorithm to denoise Biomedical images and separate mixed natural images.

## Declaration of competing interest

The authors declare that they have no known competing financial interests or personal relationships that could have appeared to influence the work reported in this paper.